\documentclass[11pt]{article}

\usepackage[final]{acl}

\usepackage{times}
\usepackage{latexsym}
\usepackage{colortbl}

\usepackage[T1]{fontenc}

\usepackage[utf8]{inputenc}

\usepackage{microtype}

\usepackage{inconsolata}

\usepackage{graphicx}
\usepackage{multirow}
\usepackage{hhline}
\usepackage{booktabs}
\usepackage{xspace}
\usepackage{makecell}
\usepackage{vcell}
\usepackage{hyperref}
\usepackage{xcolor}
\usepackage{float}

\usepackage[normalem]{ulem}
\useunder{\uline}{\ul}{}

\newcommand{\model}{AMALIA\xspace}

\newcommand{\pticoline}{\raisebox{-0.3ex}{\includegraphics[height=2ex]{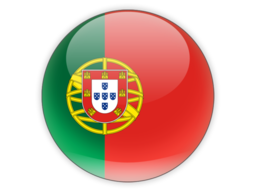}}}

\newcommand{\enicoline}{\raisebox{-0.3ex}{\includegraphics[height=2ex]{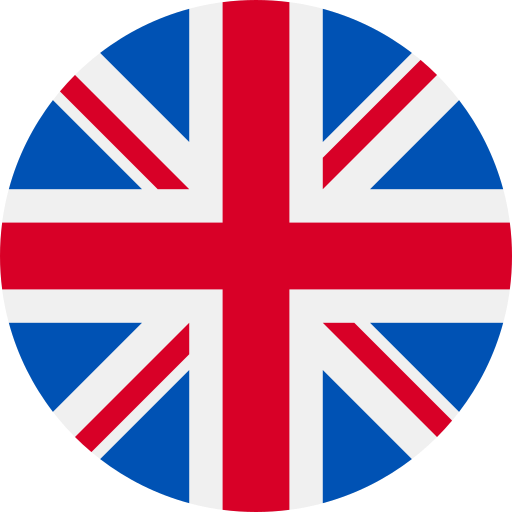}}}

\usepackage{pifont}
\newcommand{\cmark}{\ding{51}}

\title{AMALIA Technical Report: \\A Fully Open Source Large Language Model for European Portuguese}

\author{
 \textnormal{Afonso Simplício\textsuperscript{1,2},}
 \textnormal{Gonçalo Vinagre\textsuperscript{1,2},}
 \textnormal{Miguel Moura Ramos\textsuperscript{3,4},}
 \textnormal{Diogo Tavares\textsuperscript{1,2},}
 \textnormal{Rafael Ferreira\textsuperscript{1,2},}
\\
 Giuseppe Attanasio\textsuperscript{3},
 Duarte M. Alves\textsuperscript{3,4},
 Inês Calvo\textsuperscript{1},
 Inês Vieira\textsuperscript{1},
 Rui Guerra\textsuperscript{1},
 James Furtado\textsuperscript{1,2},
\\
 Beatriz Canaverde\textsuperscript{3,4},
 Iago Paulo\textsuperscript{1,2},
 Vasco Ramos\textsuperscript{1,2},
 Diogo Glória-Silva\textsuperscript{1,2},
 Miguel Faria\textsuperscript{3},
 Marcos Treviso\textsuperscript{3,4},
\\
 Daniel Gomes\textsuperscript{5},
 Pedro Gomes\textsuperscript{5},
 David Semedo\textsuperscript{1,2},
 André Martins\textsuperscript{3,4},
 João Magalhães\textsuperscript{1,2}
\vspace{6pt}
\\
 \textsuperscript{1}NOVA School of Science and Technology,
 \textsuperscript{2}NOVA LINCS, 
 \textsuperscript{3}Instituto de Telecomunicações, 
 \\
 \textsuperscript{4}Instituto Superior Técnico, Universidade de Lisboa, 
 \textsuperscript{5}Fundação para a Ciência e Tecnologia
\\
 \small{
   \textbf{Correspondence:} \href{mailto:{am.simplicio,gv.martins}@campus.fct.unl.pt}{\{am.simplicio, gv.martins\}@campus.fct.unl.pt}
 }
}

\begin{document}
\maketitle
\begin{abstract}

Despite rapid progress in open large language models (LLMs), European Portuguese (pt-PT) remains underrepresented in both training data and native evaluation, with machine-translated benchmarks likely missing the variant's linguistic and cultural nuances. We introduce \model, a fully open LLM that prioritizes pt-PT by using more high-quality pt-PT data during both the mid- and post-training stages.
To evaluate pt-PT more faithfully, we release a suite of pt-PT benchmarks that includes translated standard tasks and four new datasets targeting pt-PT generation, linguistic competence, and pt-PT/pt-BR bias.
Experiments show that \model matches strong baselines on translated benchmarks while substantially improving performance on pt-PT-specific evaluations, supporting the case for targeted training and native benchmarking for European Portuguese.\footnote{\url{https://github.com/AMALIA-LLM/AMALIA}}

\end{abstract}

\section{Introduction}
\label{sec_intro}
The rapid development of large language models (LLMs) has fundamentally transformed the field of natural language processing (NLP), with open-weight models~\cite{grattafiori2024llama, yang2025qwen3technicalreport, gemma_2025} setting new standards in tasks ranging from problem-solving to conversational AI.
However, a significant limitation of current state-of-the-art models is their overwhelming focus on English-language data~\cite{olmo2025olmo3}.
This imbalance is reflected in evaluation, where the scarcity of native benchmarks leads many researchers to rely on machine translation~\cite[e.g.,][]{son-etal-2025-kmmlu,Xuan2025MMLUProXAM}.
In turn, such MT benchmarks often fail to capture the linguistic and cultural nuances of content naturally produced by native speakers~\cite{Plaza2024SpanishAL,Wu2025TheBL}, spurring focused efforts to prioritize native, culturally specific sourcing~\cite[e.g.,][]{attanasio-etal-2024-calamita,singh-etal-2025-global}.
As a result, many European languages, including European Portuguese, are under-represented in global LLMs, limiting these technologies' ability to capture the full breadth of Europe's linguistic and cultural diversity.

This paper introduces \model, an LLM designed to address this imbalance by prioritizing European Portuguese and its cultural context during pretraining and post-training.
Furthermore, we introduce a set of benchmarks that assess the linguistic performance and cultural sensitivity of LLMs, explicitly tailored for European Portuguese.
The process of creating \model began with the collection and processing of large-scale European Portuguese data and subsequent data quality filtering.
We leveraged Arquivo.pt, the Portuguese Web Archive,\footnote{\url{https://arquivo.pt/}} as a primary source for this data.
The model was then pretrained on this data and post-trained on specific datasets for instruction following, conversational reasoning, and problem solving.

Our experiments reveal that the model is on par with other similarly-sized open models in most machine-translated benchmarks and is superior on the European Portuguese benchmarks.
This is a strong indication that LLMs can indeed capture specific traits of underrepresented language varieties, even when the number of examples is orders of magnitude smaller than the dominant language variety and the dominant language, \textit{i.e.}, Brazilian Portuguese and English, respectively.

\section{Related Work}

The development of fully open large language models has been dominated by English-centric training~\cite{olmo2025olmo3}, although recent efforts have been made in multilingual models~\cite{gonzalez2025salamandra, eurollm9btech, apertus}. Despite this, lower resource languages are still under-represented in these models, and language varieties distinction, such as  European and Brazilian Portuguese, is still a major challenge, leading to degraded performance on language-specific tasks and cultural contexts. 

To mitigate this phenomena, several efforts have addressed the lack of data and models for European Portuguese specifically. Gloria~\cite{gloria_llm} introduced an European Portuguese decoder-based LLM pretrained on a large PT-PT text corpora, and Gervasio~\cite{santos-etal-2024-advancing} fine-tuned Llama-Instruct 2~\cite{grattafiori2024llama} for pt-PT.

\section{Data Collection and Processing}
\label{sec_data_collection}

\subsection{Arquivo Data}
\label{sub_arquivo_data}

\paragraph{Collections.}
We collected publicly available data from Arquivo.pt. Following GlórIA's~\cite{gloria_llm} efforts, we gathered WARC archives comprising a selection of collections that primarily contain general web and free books, thereby improving the variety and quantity of the Portuguese data in the pretraining phase. We obtained this data by copying directly from Arquivo.pt's data center; in its raw format, it totals 195 terabytes of WARC archives, which were subsequently processed.

\paragraph{Filtering.}
We created a data processing and filtering pipeline inspired by the FineWeb2~\cite{penedo2025fineweb2pipelinescale} pipelines, using the Datatrove~\cite{penedo2024datatrove} library.
For each data collection, we begin by using URL filtering to remove any `.br' domains to reduce Brazilian Portuguese content. We also use a blacklist to remove sensitive (NSFW) content. Then, we scrape the text from the HTML files using the Trafilatura~\cite{barbaresi-2021-trafilatura} extractor and perform additional post-scraping by removing any short or duplicate lines in the text.
We then followed FineWeb heuristic filters with its parameters calibrated for European Portuguese, namely the Language Identification~\cite{Kargaran_2023}, FineWeb Quality, and Gopher Repetition and Quality filters~\cite{rae2022scalinglanguagemodelsmethods}. 
After removing personal data and fixing remaining encoding issues, we deduplicated the data using MinHash~\cite{666900}, and labeled it according to the EuroFilter quality classification model~\cite{eurollm9btech}.
Finally, we merged the data in all collections, creating three quality splits (high, medium and low), and we used the high and medium quality, totaling 5.8 billion tokens.

\paragraph{GDPR compliance.}
The data from Arquivo.pt was collected from the public Web and respects an embargo of one year. As mentioned, we removed personal data, 
namely public IPs, email addresses, and phone numbers.

\subsection{Long Context}
\label{sub_long_context}
Following \citet{ramos2026eurollm22btechnicalreport}, we reused the long documents from EuroLLM's pretraining corpus~\cite{eurollm9btech}, which had previously been truncated to $4$K tokens. 
We also collected long-context code samples from Stack-v2~\cite{lozhkov2024starcoder2stackv2}, considering only repositories with at least 500 stars and 100 forks. 
Additionally, we included synthetic data focused on improving recall and retrieval in long-context scenarios.
Concretely, starting from publicly available datasets,\footnote{Datasets available at 
\url{https://huggingface.co/datasets/lvwerra/needle-llama3-16x8k} and \url{https://huggingface.co/datasets/nanotron/needle_32k_finetuning_dataset}.} we filtered out samples longer than $32$K tokens and removed duplicates.

\subsection{Post-train Data}
\label{sub_post_train_data}
To improve \model{}'s instruction following capabilities, we built a data mixture targeting four categories: instruction following, conversational reasoning, mathematical problem-solving, and safety. This mixture comprises our own synthetic and manually curated data in both European Portuguese and English alongside publicly available datasets from Hugging Face. We only used data sources that had an open-source license associated to it.
The complete data mixture is detailed in Table~\ref{tab:sft_mix}.

Our synthetic generation primarily followed the PersonaHub~\cite{ge2025scalingsyntheticdatacreation} approach to increase diversity in instruction following and conversational scenarios, using the PersonaHub and Nemotron personas~\cite{nvidia/Nemotron-Personas-USA} to create datasets for general, instruction following, and mathematical tasks. European Portuguese data was produced via machine translation using our pt-PT MT model and Gemma 3-27B~\cite{gemma_2025}, question generation with 
few-shot prompting, and answer generation with Gemma 3-27B. This choice stems from our observation that Gemma 3-27B was the best available open model for European Portuguese, a finding later corroborated by experimental results (Section~\ref{sub_pt_pt_generation}). 
We consistently relied on larger and varied models to produce prompts, questions, and answers in order to reduce the risk of biased data sampling. 

To ensure high data quality, we applied dataset-specific filtering and formatting strategies. We removed all reasoning traces, eliminated samples containing self-referential content from other models, and utilized the \textsc{Deita} Quality Scorer~\cite{liu2023what}, which gives a score between 1 and 6 to filter out synthetic entries with quality scores below 5. Finally, we applied global deduplication, retaining only one entry per unique user prompt.

\begin{table*}
\centering
\arrayrulecolor[rgb]{0.753,0.753,0.753}
\resizebox{\linewidth}{!}{%
\begin{tabular}{cllcrl} 
\toprule
\textbf{Task} & \textbf{Dataset} & \textbf{Language} & \textbf{Proportion} & \textbf{\#Tokens} & \textbf{License / Source} \\ 
\midrule
 & Persona-PT Instruction Following & PT & 0.332\% & 10 741 619 & Apache 2.0 / Synth. Gen. \\ 
 & Persona-EN Instruction Following & EN & 0.807\% & 26 065 053 & Apache 2.0 / Synth. Gen. \\
 & Persona Nemotron Instruction Following & PT & 0.147\% & 4 765 186 & Apache 2.0 / Synth. Gen. \\
 \multirow{-4}{*}{{}\begin{tabular}[c]{@{}>{}c@{}}\textbf{Instruction}\\\textbf{Following}\end{tabular}} & Portuguese Linguistic Instructions & PT & 0.001\% & 18 884 & Apache 2.0 / Manual \\ 
\midrule
 & \model Hardcoded & PT \& EN & 0.001\% & 41 756 & Apache 2.0 / Manual \\
 & Persona-PT General & PT & 5.945\% & 192 102 099 & Apache 2.0 / Synth. Gen. \\
 & Persona Nemotron General & PT & 2.906\% & 93 916 092 & Apache 2.0 / Synth. Gen. \\
 & Wikipedia Conversations & PT & 2.336\% & 75 493 190 & Apache 2.0 / Synth. Gen. \\
 & \href{https://huggingface.co/datasets/nvidia/Nemotron-Post-Training-Dataset-v1/viewer/default/chat}{Nemotron-Post-Training-v1 - Chat Split} & Multi & 13.206\% & 426 741 052 & CC BY 4.0 / HF\\
 & \href{https://huggingface.co/datasets/nvidia/Nemotron-Post-Training-Dataset-v1/viewer/default/stem}{Nemotron-Post-Training-v1 - STEM Split} & EN & 11.616\% & 375 380 057 & CC BY 4.0 / HF\\
 & \href{https://huggingface.co/datasets/nvidia/Nemotron-Post-Training-Dataset-v2}{Nemotron-Post-Training-v2 - Chat Split} & Multi & 10.993\% & 355 252 481 & CC BY 4.0 / HF \\
 & \href{https://huggingface.co/datasets/nvidia/Nemotron-Post-Training-Dataset-v2}{Nemotron-Post-Training-v2 - STEM Split} & Multi & 4.757\% & 153 720 957 & CC BY 4.0 / HF \\
 & \href{https://huggingface.co/datasets/HuggingFaceTB/smoltalk/viewer/smol-magpie-ultra/train}{smoltalk - Smol-Magpie-Ultra (Excellent Quality)} & EN & 0.118\% & 3 809 670 & Apache 2.0 / HF \\
 & \href{https://huggingface.co/datasets/HuggingFaceTB/smoltalk/viewer/smol-summarize}{smoltalk - Smol-summarize (Translated)} & PT & 1.703\% & 55 028 254 & Apache 2.0 / HF \\
 & \href{https://huggingface.co/datasets/HuggingFaceTB/smoltalk2/viewer/SFT/smoltalk_smollm3_everyday_conversations_no_think}{smoltalk2 - everyday-conversations (Translated)} & PT & 0.023\% & 741 216 & Apache 2.0 / HF \\
 & \href{https://huggingface.co/datasets/HuggingFaceTB/smoltalk2/viewer/SFT/table_gpt_no_think}{smoltalk2 - Table-GPT} & EN & 0.317\% & 10 257 869 & MIT / HF \\
 & \href{https://huggingface.co/datasets/NousResearch/Hermes-3-Dataset}{Hermes3 Custom ST Split} & EN & 2.454\% & 79 297 721 & Apache 2.0  / HF\\
 & \href{https://huggingface.co/datasets/liaad/PTradutor}{PTradutor} & PT $\leftrightarrow$ EN & 0.740\% & 23 912 149 & MIT / HF\\
 & \href{https://huggingface.co/datasets/google/wmt24pp}{WMT'24 Multilingual Translations} & Multi → PT & 0.055\% & 1 766 630 & Apache 2.0 / HF \\
 \multirow{-16}{*}{{}\begin{tabular}[c]{@{}>{}c@{}}\textbf{\textbf{Conversational}}\\\textbf{\textbf{Reasoning}}\end{tabular}} & \href{https://huggingface.co/datasets/allenai/tulu-3-sft-olmo-2-mixture}{tulu-3-sft-olmo-2-mixture} & Multi & 18.540\% & 599 099 985 & ODC-BY-1.0. / HF \\ 
\midrule
 & Persona-PT Math & PT & 1.256\% & 40 596 577 & Apache 2.0 / Synth. Gen. \\
 & Persona Nemotron Math & PT & 2.847\% & 91 995 370 & Apache 2.0 / Synth. Gen. \\
 & \href{https://huggingface.co/datasets/nvidia/Nemotron-Post-Training-Dataset-v1/viewer/default/math}{Nemotron-Post-Training-v1 - Math Split} & EN & 12.340\% & 398 746 982 & CC BY 4.0 / HF \\
 & \href{https://huggingface.co/datasets/nvidia/Nemotron-Post-Training-Dataset-v2}{Nemotron-Post-Training-v2 - Math Split} & EN & 2.431\% & 78 563 288 & CC BY 4.0 / HF \\
 & \href{https://huggingface.co/datasets/nvidia/Nemotron-Post-Training-Dataset-v2}{Nemotron-Post-Training-v2 - Code Split} & EN & 1.656\% & 53 498 843 & CC BY 4.0 / HF \\
\multirow{-6}{*}{{}\textbf{\textbf{Mathematics}}} & \href{https://huggingface.co/datasets/microsoft/orca-math-word-problems-200k}{orca-math-word-problems-200k} & EN & 1.793\% & 57 947 703 & MIT / HF \\ 
\midrule
 & EuroBlocks Safety Samples & Multi & 0.422\% & 13 650 597 & Apache 2.0 / HF \\
\multirow{-2}{*}{{} \textbf{\textbf{\textbf{\textbf{Safety}}}}} & AI Generated Safety Data & EN & 0.258\% & 8 346 310 & Apache 2.0 / Synth. Gen. \\
\bottomrule
\end{tabular}
}
\caption{Data mixture used in the Supervised Fine-Tuning stage. Datasets were generated synthetically (Synth. Gen.), created manually (Manual) or pre-existent in HuggingFace (HF).}
\vspace{-2mm}
\label{tab:sft_mix}
\end{table*}

\paragraph{Instruction Following.}
The instruction following mixture consists of datasets created primarily via the aforementioned PersonaHub method. To address pt-PT specific nuances, we included a Portuguese Linguistic Instructions dataset, manually curated by a Portuguese linguistics expert. This dataset, comprising 200 entries, covers aspects, such as phonetics, orthography, wordplay, idiomatic expressions, and grammar classification, specific to European Portuguese.

\paragraph{Conversational Reasoning.}
This mixture provides general knowledge in a conversational format, incorporating the Persona-PT General and Persona Nemotron General datasets, and generated conversational data from Wikipedia.
We also included the \model Hardcoded dataset, providing 156 entries with self-referential knowledge about \model's development and capabilities.  

To improve system-prompt adherence~\cite{system-message-gen,system-message-robustness}, we manually selected a subset of the Hermes3 SFT dataset~\cite{teknium2024hermes3technicalreport}, retaining only entries containing custom system prompts that substantially modify the model's behavior. Furthermore, from the smoltalk's smol-magpie-ultra subset~\cite{Allal2025SmolLM2WS} we kept the entries pre-labeled as excellent quality and translated the smol-summarize subset. We also processed the smoltalk2's everyday-conversations subset~\cite{bakouch2025smollm3} by translating initial turns, generating remaining turns in European Portuguese and removing all greeting turns. We randomly selected 1M entries from the STEM split of Nemotron-Post-Training-v1~\cite{NemotronPostTrainingDatasetV1, bercovich2025llamanemotronefficientreasoningmodels}, and removed entries with noisy artifacts.
From the Nemotron-Post-Training-v2~\cite{NemotronPostTrainingDatasetV2, nvidia2025nvidianemotronnano2} chat split, following RIP~\cite{yu2025ripbettermodelssurvival}, we removed problematic instructions from the WildChat~\cite{zhao2024wildchat1mchatgptinteraction} subset.

For translation capabilities, we used the PTradutor dataset~\cite{Sousa2025} for PT-EN and EN-PT translations, by selecting bidirectional translation pairs, and the WMT24++ dataset~\cite{deutsch2025wmt24expandinglanguagecoverage}, where we crossed the XX-EN pairs with the respective PT-EN pairs to obtain XX-PT pairs and downsampled them to balance the source-target variety.

\paragraph{Mathematics.}
To increase the presence of mathematical data in European Portuguese, we generated the Persona-PT Math and Persona Nemotron Math datasets. 
To improve robustness to varying output formats, we modified the math split of Nemotron-Post-Training-v2 by removing the mandatory 
\texttt{\textbackslash boxed\{x\}} termination from 50\% of the samples, improving robustness to alternative valid output formats.

\paragraph{Safety.}
For the safety mixture, we classified every prompt of the Euroblocks SFT dataset\footnote{\url{https://huggingface.co/datasets/utter-project/EuroBlocks-SFT-Synthetic-1124}}
using Qwen3Guard~\cite{qwen_guard_3}, retaining only samples that contained unsafe or controversial prompts paired with the dataset's safe responses.
Additionally, we created an English safety dataset using DeepSeek-V3.2-Exp~\cite{deepseekai2024deepseekv32} that covers hierarchically organized sensitive topics and adversarial strategies.

\section{Model Development}
\label{sec_model_development}
To build the base \model{} model, we modified the final pretraining phase of EuroLLM-9B~\cite{eurollm9btech} to improve its language modeling capabilities for pt-PT, and carried a supervised fine-tuning phase for instruction following and conversational reasoning.

\subsection{Pretraining}
\label{sub_pretraining}

The data sources used in \model's pretraining consist of four sources. First, the original EuroLLM training mixture, which contains 40B tokens from multiple European languages---including a small portion of mathematics and code---and is largely composed of short documents rather than long sequences.
To better support longer contexts, we expanded it with 60B tokens with improved coverage of the code domain, and further included 1.4B synthetically generated long-context tokens. This strategy balances shorter and longer sequences, which improves performance on long-context tasks while maintaining model quality on short-context tasks~\cite{gao2025prolong}. Additionally, to further improve European Portuguese capabilities, we included 5.8B tokens from the Arquivo.pt dataset (\S\ref{sub_arquivo_data}).
We largely follow EuroLLM-9B hyperparameters, with two changes: we extend the max sequence length from 4K to 32K tokens and apply RoPE scaling~\cite{gao2025prolong}, increasing $\theta$ from 10,000 to 1,000,000. Training took 80 hours on 256 NVIDIA H100 GPUs.

\subsection{Supervised Fine-Tuning}
\label{sub_sft}

We fine-tuned \model{} to improve its instruction following and conversational capabilities using the data mixture described in Table~\ref{tab:sft_mix}. Training was conducted over 14K steps (approximately 4.25 epochs), using the AdamW optimizer~\citep{loshchilov2017decoupled} with a learning rate of $10^{-5}$, a cosine learning rate scheduler with a warmup ratio of 0.03, weight decay of 0.01, and \texttt{bfloat16} mixed precision. The final checkpoint was selected based on the performance on the validation set. We trained for 76 hours on 64 NVIDIA H100 GPUs.

\subsection{Preference Training}
\label{sub_preference_training}

For preference training, we used Direct Preference Optimization (DPO)~\cite{rafailov2024directpreferenceoptimizationlanguage}, starting by following an approach inspired by OLMo3~\cite{olmo2025olmo3}, using 200K prompts sampled from the SFT dataset. We experimented with generating responses using several models; the best results were obtained by sampling 32 candidate responses from our own SFT model for each prompt, scoring them with ArmoRM~\citep{armo_rm}, and selecting the highest- and lowest-reward responses as the chosen and rejected answers, respectively. 

While this strategy led to overall performance improvements, it resulted in declines in mathematical reasoning and instruction following.
We hypothesize that these declines stem from the reward model favoring responses that do not strictly adhere to task constraints.
To mitigate these issues, we adopted domain-specific data construction strategies. For our persona datasets, we used the original answers as the chosen responses and \model-SFT generated answers as the rejected ones. For mathematical reasoning, we found that generating chosen responses with Qwen 3-32B~\cite{yang2025qwen3technicalreport} and rejected responses with Qwen 3-0.6B~\cite{yang2025qwen3technicalreport} yielded the best performance. We also observed a tendency for a decline in some Portuguese-specific capabilities; therefore, we decided to add more Portuguese data to the mix, adding subsets of the Persona-Nemotron datasets, using 100K, 50K, and 30K entries from the General, Math, and Instruction Following sets, respectively, with the rejected responses generated by Qwen 3-0.6B~\cite{yang2025qwen3technicalreport}.

\begin{table*}[tb]
\centering
\small
\resizebox{\linewidth}{!}{%
\begin{tabular}{@{}llccccc@{}}
\toprule
\textbf{Dataset}             & \textbf{Category}     & \textbf{\begin{tabular}[c]{@{}c@{}}Source  \\ Language\end{tabular}}       & \textbf{CoT} & \textbf{\begin{tabular}[c]{@{}c@{}}\#Shots\end{tabular}} & \textbf{Metric} & \textbf{Reference}        \\ \midrule
ALBA           & PT Grammar            & \pticoline            &             & 0                        & LLM-Judge & New                       \\
PT Completions & pt-PT Generation         & \pticoline            &              & 0                        & Accuracy & New                       \\
PT Exams       & General Knowledge     & \pticoline            & \cmark   & 0                        & Accuracy & New                       \\
PT Exams Open Questions      & General Knowledge     & \pticoline            & \cmark   & 0                        & LLM-Judge & New                       \\
P3B3                 & pt-PT/BR Generation     & \pticoline            &              & 0                        & pt-PT Level & New        \\
FRMT                         & Translation           & \pticoline &              & 5                        & chrF &~\cite{frmt}                 \\ 
ARC-C                        & Commonsense Reasoning & \enicoline            &              & 0                        & Accuracy &~\cite{thellmann2024towards} \\
GSM8K                        & Mathematics           & \enicoline            & \cmark       & 8                        & Exact Match &~\cite{thellmann2024towards} \\
MMLU                         & General Knowledge     & \enicoline            &              & 0                        & Accuracy &~\cite{thellmann2024towards} \\
TruthfulQA                   & NLU                   & \enicoline            &              & 6                        & Accuracy &~\cite{thellmann2024towards} \\ 
PIQA              & Commonsense Reasoning & \enicoline            &              & 0                        & Accuracy & New (MT)                    \\
SIQA              & Commonsense Reasoning & \enicoline            &              & 0                        & Accuracy & New (MT)                    \\
IFEval            & Instruction Following & \enicoline            &              & 0                        & Prompt level strict& New (MT)                    \\
BBH               & Reasoning             & \enicoline            & \cmark       & 3                        & Exact Match & New (MT)                    \\ 
Simple Safety Tests              & Safety & \enicoline            &              & 0                        & ASR & New (MT)                    \\
XSTest              & Safety & \enicoline            &              & 0                        & ASR & New (MT)                    \\
Multilingual ADV bench              & Safety & \enicoline            &              & 0                        & ASR & \href{https://huggingface.co/datasets/simonycl/multilingual_advbench}{simonycl/multilingual\_advbench}                    \\
\bottomrule
\end{tabular}%
}
\caption{Summary of datasets used for evaluation. \textbf{New} denotes novel datasets created for pt-PT evaluation. \textbf{New (MT)} are datasets newly translated into pt-PT via our pt-PT MT model. Note: Some BBH tasks were removed from pt-PT evaluation as they do not translate correctly to Portuguese (e.g. hyperbaton task).}
\label{tab_eval_datasets_list}
\vspace{-2mm}
\end{table*}

To further improve general capabilities and explicitly target weaknesses in math and instruction following, we incorporated additional preference datasets: UltraFeedback~\cite{cui2023ultrafeedback}, OpenAssistant 2~\cite{köpf2023openassistantconversationsdemocratizing}, Abbey4799/Complex-Instructions-DPO,\footnote{\url{https://huggingface.co/datasets/Abbey4799/Complex-Instructions-DPO}} and kira/math-dpo.\footnote{\url{https://huggingface.co/datasets/kira/math-dpo}} To improve model safety, we also included Egida-DPO-Meta-LLaMa-3.1-70B-Instruct~\cite{garciagasulla2025efficientsafetyretrofittingjailbreaking}, HarmfulQA~\cite{bhardwaj2023redteaming}, and a safety dataset tailored to the Portuguese cultural context.
In total, this mixture comprised 478K preference pairs.

We trained for one epoch with a batch size of 128, a learning rate of $10^{-6}$ under a linear scheduler, and $\beta=0.1$, 
taking 12 hours on 64 NVIDIA H100 GPUs.

\section{Experimental Setup}
\label{sec_experimental_setup}

Following other fully open LLMs~\citep{OLMo20242O2,Allal2025SmolLM2WS},
we evaluate instruct and preference \model variants on a suite of diverse and challenging benchmarks.

\subsection{Model Baselines}
\label{sub_baselines}

We compare \model to a range of instruction-tuned models of comparable size, including both open-weight and fully open-source models. Among the open-weight models, we consider Mistral-7B~\citep{mistral}; Ministral-8B~\cite{ministral8b2024}; Llama 3.1-8B~\citep{grattafiori2024llama}; Gemma 2-9B~\cite{gemma2}; Gemma 3-12B~\cite{gemma_2025}, all designed for broad cross-lingual generalization; and the Qwen family, comprising Qwen 2.5-7B~\citep{qwen2.5} and Qwen 3-8B~\cite{yang2025qwen3technicalreport}, which are known for strong multilingual and reasoning capabilities.
For fully open-source models that release both code and data, we evaluate OLMo 2-7B~\cite{OLMo20242O2}, a strong English-only baseline; Salamandra-7B~\citep{gonzalez2025salamandra}, focused on Iberian languages; EuroLLM-9B~\citep{eurollm9btech}, tailored to European languages; and Apertus-8B~\citep{apertus}, trained on hundreds of languages.
Additionally, we include Gervasio-8B,\footnote{\url{https://huggingface.co/PORTULAN/gervasio-8b-portuguese-ptpt-decoder}}, a Llama-Instruct 3.1-8B model, fine-tuned for pt-PT.

\subsection{PT Benchmark Collection}
\label{sub_pt_benchmark_collection}

Benchmarks were selected according to the following criteria: (1) originally authored in European Portuguese (pt-PT) or Brazilian Portuguese (pt-BR), (2) translated by humans, and (3) existing machine-translated datasets. To further expand coverage of evaluation capabilities, we additionally machine-translated benchmarks lacking a Portuguese variant and compiled new corpora to assess pt-PT relevant skills.
In all evaluations, every prompt component, including questions, answers, and instructions, were presented in pt-PT.

\subsection{pt-PT Benchmark Collection}
\label{sub_new_datasets}

Developing datasets in pt-PT requires special attention to cultural and linguistic distinctions from pt-BR. To address these differences, we introduce four new datasets designed to support benchmarking and evaluation for pt-PT language models.

\paragraph{PT-PT Completions (PT-C).}
This task evaluates a model's ability to complete sentences in European Portuguese (pt-PT). Each sentence has a blank and two options: one in pt-PT and one in pt-BR, where models should select the first option by default. The dataset contains 70 manually curated examples highlighting pt-PT/pt-BR differences, e.g., \textit{``Vou à estação de \_\_\_\_ para comprar um bilhete.''}, with options (A) ``comboios''---a pt-PT term---and (B) ``trem''---a pt-BR term.

\paragraph{PT Exams (PT-E)~\cite{tavares-etal-2026-pheb}.}
We extract questions and answers from the official Portuguese national high school exams.\footnote{\url{https://github.com/AMALIA-LLM/pheb}}
The dataset comprises 1.8K multiple-choice and 1.8K open questions from 2006–2023, over six subjects: Mathematics, Portuguese, History, Geography, Biology/Geology, and Philosophy. All questions were written and curated by educators to ensure correctness and alignment with the national curriculum. These questions assess factual knowledge, reasoning skills, and language comprehension.

\paragraph{ALBA~\cite{alba}.} 
Automated Linguistics Benchmark for baseline Assessment (ALBA) is a pt-PT linguistics benchmark designed to address gaps in existing language evaluation resources. 
Developed manually by domain experts from our team, it covers eight categories: language variety, culture-bound semantics, discourse analysis, wordplay, syntax, morphology, lexicology, and phonetics \& phonology.
The dataset comprises 800 questions (100 per category), of which 30 are paired with three graded reference answers to enable validation against human judgment.

\paragraph{P3B3.\protect\footnote{\url{https://github.com/AMALIA-LLM/p3b3-benchmark}}} There are many nuances to consider when comparing pt-PT to pt-BR, particularly with respect to grammatical structures and lexical variation. To evaluate potential model bias toward either variant, we construct the pt-PT--pt-BR Bias Benchmark (P3B3), comprising 200 multi-turn conversational prompts, each with 2 to 5 turns. The prompts are variant-agnostic and designed to provoke a variant-specific response. At each turn, the model generates a response using the accumulated dialogue history. We then use automated methods to assess whether responses align more with pt-PT or pt-BR.

\subsection{Benchmark Selection}
\label{sub_benchmark_distribution}

We evaluate \model{} on a diverse set of tasks following~\citet{OLMo20242O2, Allal2025SmolLM2WS}.
Specifically, we include ARC Challenge~\citep[ARC-C;][]{arc_c}, MMLU~\citep{mmlu}, TruthfulQA~\citep[TQA;][]{lin2022truthfulqa}, PiQA~\citep{piqa}, and SiQA~\citep{siqa}. 
Together, these tasks assess a broad spectrum of capabilities, including commonsense reasoning, scientific and world knowledge, reading comprehension, question answering, and cloze-style completion.
Additionally, we test models on more complex instruction following datasets, such as BigBenchHard~\citep[BBH;][]{srivastava2023bigbench}, IFEval~\citep[IFE;][]{zhou2023ifeval}, along with GSM8k~\citep{cobbe2021gsm8k} for math reasoning.

To more directly evaluate Portuguese-specific performance, we extend this suite with benchmarks designed to test linguistic and domain knowledge in Portuguese.
These include the pt-PT translation dataset FRMT~\citep{frmt}, and our newly created pt-PT–specific datasets from Section~\ref{sub_new_datasets}. 

We also include safety-focused benchmarks. Specifically, we evaluate on Simple Safety Tests~\citep[SST;][]{vidgen2023simplesafetytests}, a corpus of unsafe prompts spanning five harm categories; the unsafe subset of XSTest~\citep[XST;][]{rottger2024xstest},
and the Multilingual ADV Bench~\citep[MAD;][]{advbench}, which contains unsafe prompts covering a broad range of adversarial attack types.

\begin{table*}[t]
\centering
\small
\setlength{\tabcolsep}{3.5pt}
\begin{tabular}{lccccccccccc}
\toprule
\textbf{Model}                 & \textbf{ARC-C} & \textbf{MMLU} & \textbf{TQA}  & \textbf{GSM8K} & \textbf{IFEval} & \textbf{PT-E (CoT)} & \textbf{PT-C} & \textbf{PiQA} & \textbf{SiQA} & \textbf{FRMT} & \textbf{BBH}  \\ \midrule
\multicolumn{2}{l}{\textit{Fully open models}}  &               &               &                &                 &                     &               &               &               &               &               \\
OLMo 2-7B                      & 47.8           & 42.6          & 50.8          & 51.6           & 50.6            & 45.0                & 32.9          & 61.0          & 41.9          & 61.0          & 37.0          \\
Salamandra-7B                  & 52.2           & 47.1          & 52.6          & 18.7           & 24.2            & 37.7                & 44.3          & 70.4          & 44.8          & 65.6          & 36.3          \\
EuroLLM-9B                     & 71.2           & 52.7          & 54.9          & 52.5           & 49.4            & 55.2                & 58.6          & 71.7          & 40.8          & 70.4          & 47.6          \\
Apertus-8B                     & 68.3           & 57.1          & 57.9          & 48.8           & 58.6            & 55.7                & 40.0          & \textbf{73.4} & 43.2          & 67.2          & 48.3          \\
\model-9B-SFT   & 77.9           & 55.4          & 57.0          & {\ul 58.8}     & 56.7            & 63.1                & \textbf{71.4} & 71.7          & 44.7          & \textbf{70.5} & {\ul 50.3}    \\
\model-9B-DPO   & {\ul 78.9}     & {\ul 58.8}    & \textbf{63.5} & 52.4           & {\ul 61.6}      & {\ul 68.4}          & 67.1          & 72.5          & \textbf{46.3} & 70.2          & 47.3          \\ \midrule
\multicolumn{2}{l}{\textit{Open weight models}} &               &               &                &                 &                     &               &               &               &               &               \\
Llama 3.1-8B                   & 75.8           & 58.5          & 58.5          & 68.0           & 55.1            & 66.7                & 38.6          & 68.5          & 43.4          & 66.3          & 68.5          \\
Gervasio-8B                    & 79.0           & 59.8          & 56.8          & 67.3           & 55.6            & 66.0                & 30.0          & 69.1          & 44.8          & 66.3          & 68.1          \\
Gemma 2-9B                     & 85.6           & 65.8          & 62.7          & 73.1           & 47.7            & 70.9                & 34.3          & 70.3          & 41.0          & 66.5          & 50.8          \\
Gemma 3-12B                    & 86.9           & \textbf{68.0} & 63.2          & \textbf{79.2}  & 71.7            & 81.3                & 21.4          & 69.0          & 43.8          & 69.5          & \textbf{79.3} \\
Qwen 2.5-7B                    & 83.7           & 65.7          & 63.2          & 74.0           & 61.6            & 73.7                & 40.0          & 66.6          & 40.7          & 63.9          & 56.5          \\
Qwen 3-8B                      & \textbf{87.4}  & 64.9          & 57.2          & 77.6           & \textbf{72.1}   & \textbf{82.8}       & 28.6          & 64.7          & 41.2          & 65.1          & 23.3          \\
Mistral-7B                     & 61.8           & 51.0          & 61.1          & 38.3           & 39.7            & 53.6                & 34.3          & 66.9          & 40.9          & 63.3          & 52.1          \\
Ministral-8B                   & 73.6           & 55.5          & 56.0          & 66.9           & 44.0            & 64.9                & 40.0          & 70.7          & 42.3          & 66.1          & 59.1          \\ \bottomrule
\end{tabular}
\caption{Results of instruction-tuned variants. Bold represents best model and underline best fully open model.}
\label{tab_instruction_tuned_results}
\end{table*}

\subsection{Benchmark Configuration}
\label{sub_benchmark_config}

We performed evaluations using a combination of LM-Eval-Harness,\footnote{\url{https://github.com/EleutherAI/lm-evaluation-harness}} an evaluation framework that supports custom task development, and custom scripts for LLM-as-a-Judge evaluation.
Inference was conducted on a single NVIDIA H100 GPU, with models loaded in \texttt{bf16} precision and served through vLLM~\cite{vllm}, using automatic batch sizing and a maximum context length of 4096 tokens.
Whenever available, we used the official task generation parameters provided by each benchmark and used greedy decoding when such parameters were not specified.
Following OLMES standards~\citep{gu-etal-2025-olmes}, tasks are formatted either as multiple-choice or chain-of-thought prompting. We evaluate models in zero- and few-shot settings and report the metric associated with each dataset. 

Table~\ref{tab_eval_datasets_list} summarizes the full benchmark configuration. For each dataset, we include its source and translation method. In addition to the newly constructed datasets tailored for pt-PT evaluation, several datasets were automatically translated using a pt-PT translation model, which is explicitly optimized for pt-PT translation, since models tend to translate to pt-BR.

\section{Results and Discussion}
\label{sec_results}

\subsection{Instruct Model Evaluation}
\label{sub_instruct_model_eval}

We show the performance of instruction-tuned variants of \model{} on multiple tasks in Table~\ref{tab_instruction_tuned_results}.
Overall, \model{} achieves state-of-the-art performance among fully open models of comparable size across most benchmarks. The SFT variant leads on four benchmarks, exhibiting particularly strong performance in Portuguese language understanding and grade school mathematics. Building on this foundation, the preference-tuned (DPO) version outperforms all other fully open models on six tasks, with especially notable improvements on PT-E, reflecting a strong grasp of the Portuguese school curriculum.
In addition, \model's performance on ARC-C, MMLU, TQA, PiQA, SiQA indicate that it acquired good general and common-sense knowledge, and IFEval results demonstrate the models flexibility across varied instruction-following scenarios.

\subsection{PT-Exams: Open Questions}
\label{sub_pt_exams_long_form}

After evaluating the model's knowledge using the MCQ from PT-Exams (Section~\ref{sub_new_datasets}), we further assess its ability to comprehend and produce written pt-PT. To this end, we again use the national Portuguese exams, but now focus on the open-ended questions for the Portuguese subject, totaling 336 questions.
Unlike MCQs, these questions do not have a single correct answer. Instead, the official exams provide correction rubrics that specify both the expected content and the quality of the written response. We extract the exam questions and evaluation criteria for the Portuguese subject and use Gemini 2.5 Pro to assess model-generated answers according to these rubrics. This evaluator was selected due to its strong agreement with human judgments~\cite{tavares-etal-2026-pheb}.

Table~\ref{pt_exams_long_form} shows that \model{} achieves the highest performance among fully open models with significant increases with DPO, demonstrating strong Portuguese text understanding and clear, grammatically correct language generation.

\subsection{ALBA: pt-PT Linguistics Evaluation}
\label{sub_alba_eval}

We evaluate pt-PT linguistic competence using the ALBA dataset (Section~\ref{sub_new_datasets}), which covers a broad range of linguistic phenomena. Evaluation is performed using Gemini 2.5 Pro as an automatic judge, producing scores on a 1–5 scale that are rescaled to a 0–100 range.
Results in Table~\ref{pt_exams_long_form} show that \model{}-DPO achieves the strongest performance among fully open models. 

Drilling down on individual categories~\cite{alba}, we observed that \model-DPO achieves the best results in both Lexicology and Culture-bound Semantics, surpassing Gemma 3 and highlighting its above-average understanding of pt-PT–specific linguistic skills.
In contrast, we found that Phonology and Wordplay emerge as the most challenging categories,
reflecting the intrinsic difficulty of these more complex and nuanced tasks. 
Despite these variations, \model{}-DPO showcases a solid performance across a broad range of pt-PT linguistic traits.

\begin{table}[tb]
\centering
\small
\setlength{\tabcolsep}{4pt}
\begin{tabular}{lccc}
\toprule
\textbf{Model} & \textbf{\begin{tabular}[c]{@{}c@{}}PT-Exams \\Open Questions\end{tabular}} & \textbf{ALBA}    & \textbf{\begin{tabular}[c]{@{}c@{}}P3B3 \end{tabular}} \\ \midrule
\textit{Fully open models}  &               &               &      \\
OLMo 2-7B                   & 43.0          & 16.9          & 18.6 \\
Salamandra-7B               & 34.1          & 27.4          & 42.7 \\
EuroLLM-9B                  & 56.1          & 38.5          & 70.5 \\
Apertus-8B                  & 54.7          & 38.7          & 28.1 \\
\model-9B-SFT               & 62.0          & 37.1          & 91.3 \\
\model-9B-DPO  & \underline{66.0}                                              & \underline{43.6} & \textbf{95.9}                                                   \\ \midrule
\textit{Open weight models} &               &               &      \\
Llama 3.1-8B                & 53.8          & 31.3          & 27.8 \\
Gervasio-8B                 & 53.2          & 31.1          & 24.7 \\
Qwen 2.5-7B                 & 56.8          & 31.0          & 20.0 \\
Qwen 3-8B                   & \textbf{77.3} & 49.8          & 18.9 \\
Gemma 2-9B                  & 69.7          & 41.1          & 72.1 \\
Gemma 3-12B                 & 76.6          & \textbf{51.1} & 88.3 \\
Mistral-7B                  & 44.6          & 21.7          & 19.2 \\
Ministral-8B                & 62.0          & 35.6          & 22.1 \\ \bottomrule
\end{tabular}%
\caption{Language generation results of instruction-tuned variants on the newly created pt-PT benchmarks.}
\label{pt_exams_long_form}
\end{table}

\subsection{P3B3: Model Bias and pt-PT Generation}
\label{sub_pt_pt_generation}

We assess the models' ability to generate pt-PT using the P3B3 dataset (Section~\ref{sub_new_datasets}) by explicitly instructing the model to respond in pt-PT. This setup allows us to evaluate both controllability and fidelity to the specified language variant.
For automatic evaluation, we use Gemini 2.5-Pro with a specialized prompt that analyzes linguistic features, produces a short rationale, and assigns a score, with scores showing strong agreement with human judgments.

Table~\ref{pt_exams_long_form} shows that most models exhibit a strong bias toward pt-BR, reflecting its higher-resource status. Even though models were explicitly prompted to use pt-PT, only \model, EuroLLM, and Gemma are able to consistently write in pt-PT. 
In specific, both versions of \model{} achieve the highest pt-PT scores on P3B3, indicating a strong ability produce outputs that conform to salient linguistic features of European Portuguese.

\subsection{Safety Results}
\label{sub_safety}

To evaluate safety, we employ a set of safety benchmarks covering both commonsense safety and jailbreak scenarios. We report the attack success rate (ASR) as assessed by Qwen3Guard-8B~\citep{qwen_guard_3}. An attack is considered successful if the classifier labels the model's response as either ``unsafe'' or ``controversial''.  

As shown in Table~\ref{tab_safety_results}, all \model{} variants exhibit strong safety performance, achieving consistently low ASR across benchmarks. Notably, \model-DPO achieves the lowest ASR among all models on two datasets and delivers the best performance among fully open models on SST, indicating a strong ability to resist jailbreak attempts. Furthermore, our DPO step further strengthened the safety of the SFT model, cutting its ASR by more than half across the three reported benchmarks.

\begin{table}[tb]
\centering
\small
\setlength{\tabcolsep}{4pt}
\begin{tabular}{@{}lrrr@{}}
\toprule
\textbf{Model}  & \textbf{ADV-Bench} $\downarrow$ & \textbf{SST} $\downarrow$ & \textbf{XSTest} $\downarrow$ \\ \midrule
\multicolumn{2}{@{}l}{\textit{Fully open models}} \\
OLMo 2-7B             & 6.9  & 20.4         & 17.4 \\
Salamandra-7B         & 6.6  & 18.4         & 23.4 \\
EuroLLM-9B   & 1.4  & 3.4          & 1.9  \\
Apertus-8B   & 2.7  & 4.4          & 3.9  \\
\model-9B-SFT             & 1.9  & 3.4          & 1.9  \\
\model-9B-DPO               & \textbf{0.8}                & \underline{1.4}           & \textbf{0.9}              \\ \midrule
\multicolumn{2}{@{}l}{\textit{Open weight models}} \\
Llama 3.1-8B & 10.8 & 7.4          & 2.9  \\
Gervasio-8B & 12.9  & 9.4          & 3.9  \\
Qwen 2.5-7B           & 1.8  & 3.4          & 2.4  \\
Qwen 3-8B             & 2.1  & 4.4          & 4.4  \\
Gemma 2-9B           & 1.0  & \textbf{0.4} & 1.4  \\
Gemma 3-12B           & 2.9  & 2.4 & 3.4  \\
Mistral-7B            & 34.4 & 19.4         & 13.4 \\
Ministral-8B          & 18.3 & 8.4          & 9.9  \\ \bottomrule
\end{tabular}%
\caption{ASR results of instruction-tuned variants on safety benchmarks assessed by Qwen3Guard-8B.}
\label{tab_safety_results}
\end{table}

\section{Conclusions}
\label{sec_conclusions}

This paper presents \model{}, an LLM that prioritizes the European Portuguese language and its cultural context. 
\model{} leverages Arquivo.pt data and post-training data specifically curated for European Portuguese, and was trained in a three-stage process: pretraining, supervised fine-tuning, and preference tuning.
To address the scarcity of suitable pt-PT benchmarks, we developed dedicated benchmarks tailored to European Portuguese, designed to capture linguistic and cultural nuances. Furthermore, we translated existing benchmarks from EN$\rightarrow$pt-PT using a dedicated translation model, and then complement the suite by introducing a national high-school pt-PT exams benchmark. 
Our results show that \model{} outperforms previously released fully-open models on European Portuguese.
In language understanding and reasoning, the model achieves state-of-the-art or comparable performance, while on language generation tasks, it excels in the quality and fluency of pt-PT text. Safety evaluations further indicate that the model is aligned with state-of-the-art standards.

\section{Future Work}

Although \model{} represents a significant step toward better fully-open European Portuguese language models, there are areas that require a continued development and improvement. One of the main challenges is the scarcity of pt-PT datasets for both training and evaluation, making it difficult to improve and assess capabilities on Portuguese-specific cultural knowledge and tasks. Another challenge is limited computational resources for pre-training, where compared to larger models we are limited to a few Trillion tokens.

We plan to continue developing new training data mixtures aimed at improving reasoning capabilities in pt-PT and pt-BR, to extend the context window, and to experiment with additional post-training approaches such as curriculum learning and reinforcement learning with verifiable rewards. 
We also aim to further enhance \model{}'s Portuguese cultural knowledge by developing targeted training data. 
Expanding multilingual capabilities and introducing tool-calling functionality are other priorities for future iterations. Finally, we aim to continuously expand our benchmark suite with new pt-PT and pt-BR evaluations covering core tasks such as instruction following, mathematical reasoning, and linguistic competence, alongside assessments of cultural and regional knowledge.

\section*{Acknowledgments}
This work was supported by the AMALIA project under Measure RE-C05-i08 of the Portuguese national Programa de Recuperação e Resiliência. We also acknowledge the support of Fundação para a Ciência e Tecnologia (FCT) and the NOVA LINCS project (UID/04516/2025). Finally, we thank the Barcelona Supercomputing Center (BSC) for providing the computational resources that made this work possible.
The IT team is supported by the Portuguese Recovery and Resilience Plan through project C645008882-00000055 (Center for Responsible AI), by the project DECOLLAGE (ERC-2022-CoG 101088763), and by FCT/MECI through national funds and when applicable co-funded EU funds under UID/50008: Instituto de Telecomunicações.

\bibliography{custom}

@article{Wu2025TheBL,
  title={The Bitter Lesson Learned from 2,000+ Multilingual Benchmarks},
  author={Minghao Wu and Weixuan Wang and Sinuo Liu and Huifeng Yin and Xintong Wang and Yu Zhao and Chenyang Lyu and Longyue Wang and Weihua Luo and Kaifu Zhang},
  journal={ArXiv},
  year={2025},
  volume={abs/2504.15521},
  url={https://api.semanticscholar.org/CorpusID:277993848}
}

@article{Xuan2025MMLUProXAM,
  title={MMLU-ProX: A Multilingual Benchmark for Advanced Large Language Model Evaluation},
  author={Weihao Xuan and Rui Yang and Heli Qi and Qingcheng Zeng and Yunze Xiao and Yun Xing and Junjue Wang and Huitao Li and Xin Li and Kunyu Yu and Nan Liu and Qingyu Chen and Douglas Teodoro and Edison Marrese-Taylor and Shijian Lu and Yusuke Iwasawa and Yutaka Matsuo and Irene Li},
  journal={ArXiv},
  year={2025},
  volume={abs/2503.10497},
  url={https://api.semanticscholar.org/CorpusID:276961588}
}

@article{Plaza2024SpanishAL,
  title={Spanish and LLM Benchmarks: Is MMLU Lost in Translation?},
  author={Irene Plaza and Nina Melero and Cristina del Pozo and Javier Conde and Pedro Reviriego and Marina Mayor-Rocher and Mar{\'i}a Grandury},
  journal={GACLM},
  year={2024},
  pages={104-108},
  url={https://api.semanticscholar.org/CorpusID:270737791}
}

@article{OLMo20242O2,
  title={2 OLMo 2 Furious},
  author={Team OLMo and Pete Walsh and Luca Soldaini and Dirk Groeneveld and Kyle Lo and Shane Arora and Akshita Bhagia and Yuling Gu and Shengyi Huang and Matt Jordan and Nathan Lambert and Dustin Schwenk and Oyvind Tafjord and Taira Anderson and David Atkinson and Faeze Brahman and Christopher Clark and Pradeep Dasigi and Nouha Dziri and Michal Guerquin and Hamish Ivison and Pang Wei Koh and Jiacheng Liu and Saumya Malik and William Merrill and Lester James Validad Miranda and Jacob Daniel Morrison and Tyler C. Murray and Crystal Nam and Valentina Pyatkin and Aman Rangapur and Michael Schmitz and Sam Skjonsberg and David Wadden and Christopher Wilhelm and Michael Wilson and Luke S. Zettlemoyer and Ali Farhadi and Noah A. Smith and Hanna Hajishirzi},
  journal={ArXiv},
  year={2024},
  volume={abs/2501.00656},
  url={https://api.semanticscholar.org/CorpusID:275213098}
}

@article{thellmann2024towards,
  author       = {Klaudia Thellmann and
                  Bernhard Stadler and
                  Michael Fromm and
                  Jasper Schulze Buschhoff and
                  Alex Jude and
                  Fabio Barth and
                  Johannes Leveling and
                  Nicolas Flores{-}Herr and
                  Joachim K{\"{o}}hler and
                  Ren{\'{e}} J{\"{a}}kel and
                  Mehdi Ali},
  title        = {Towards Multilingual {LLM} Evaluation for European Languages},
  journal      = {CoRR},
  volume       = {abs/2410.08928},
  year         = {2024},
  url          = {https://doi.org/10.48550/arXiv.2410.08928},
  doi          = {10.48550/ARXIV.2410.08928},
  eprinttype    = {arXiv},
  eprint       = {2410.08928},
  bibsource    = {dblp computer science bibliography, https://dblp.org}
}

@misc{ramos2026eurollm22btechnicalreport,
      title={EuroLLM-22B: Technical Report}, 
      author={Miguel Moura Ramos and Duarte M. Alves and Hippolyte Gisserot-Boukhlef and João Alves and Pedro Henrique Martins and Patrick Fernandes and José Pombal and Nuno M. Guerreiro and Ricardo Rei and Nicolas Boizard and Amin Farajian and Mateusz Klimaszewski and José G. C. de Souza and Barry Haddow and François Yvon and Pierre Colombo and Alexandra Birch and André F. T. Martins},
      year={2026},
      eprint={2602.05879},
      archivePrefix={arXiv},
      primaryClass={cs.CL},
      url={https://arxiv.org/abs/2602.05879}, 
}

@article{grattafiori2024llama,
  title={The Llama 3 Herd of Models},
  author={Dubey, Abhimanyu and Jauhri, Abhinav and Pandey, Abhinav and Kadian, Abhishek and Al-Dahle, Ahmad and Letman, Aiesha and Mathur, Akhil and Schelten, Alan and Yang, Amy and Fan, Angela and others},
  journal={arXiv preprint arXiv:2407.21783},
  url={https://arxiv.org/abs/2407.21783},
  year={2024}
}

@article{gonzalez2025salamandra,
  title={Salamandra technical report},
  author={Gonzalez-Agirre, Aitor and P{\`a}mies, Marc and Llop, Joan and Baucells, Irene and Da Dalt, Severino and Tamayo, Daniel and Saiz, Jos{\'e} Javier and Espu{\~n}a, Ferran and Prats, Jaume and Aula-Blasco, Javier and others},
  journal={arXiv preprint arXiv:2502.08489},
  url={https://arxiv.org/abs/2502.08489},
  year={2025}
}

@article{Allal2025SmolLM2WS,
  title={SmolLM2: When Smol Goes Big - Data-Centric Training of a Small Language Model},
  author={Loubna Ben Allal and Anton Lozhkov and Elie Bakouch and Gabriel Mart'in Bl'azquez and Guilherme Penedo and Lewis Tunstall and Andr{\'e}s Marafioti and Hynek Kydl'ivcek and Agust'in Piqueres Lajar'in and Vaibhav Srivastav and Joshua Lochner and Caleb Fahlgren and Xuan-Son Nguyen and Cl{\'e}mentine Fourrier and Ben Burtenshaw and Hugo Larcher and Haojun Zhao and Cyril Zakka and Mathieu Morlon and Colin Raffel and Leandro von Werra and Thomas Wolf},
  journal={ArXiv},
  year={2025},
  volume={abs/2502.02737},
  url={https://api.semanticscholar.org/CorpusID:276116722}
}

@inproceedings{armo_rm,
  author       = {Haoxiang Wang and
                  Wei Xiong and
                  Tengyang Xie and
                  Han Zhao and
                  Tong Zhang},
  title        = {Interpretable Preferences via Multi-Objective Reward Modeling and
                  Mixture-of-Experts},
  booktitle    = {Findings of the ACL: {EMNLP}
                  2024},
  pages        = {10582--10592},
  publisher    = {ACL},
  year         = {2024},
  url          = {https://doi.org/10.18653/v1/2024.findings-emnlp.620},
  doi          = {10.18653/V1/2024.FINDINGS-EMNLP.620},
  bibsource    = {dblp computer science bibliography, https://dblp.org}
}

@misc{ge2025scalingsyntheticdatacreation,
      title={Scaling Synthetic Data Creation with 1,000,000,000 Personas}, 
      author={Tao Ge and Xin Chan and Xiaoyang Wang and Dian Yu and Haitao Mi and Dong Yu},
      year={2025},
      eprint={2406.20094},
      archivePrefix={arXiv},
      primaryClass={cs.CL},
      url={https://arxiv.org/abs/2406.20094}, 
}

@misc{penedo2025fineweb2pipelinescale,
      title={FineWeb2: One Pipeline to Scale Them All -- Adapting Pre-Training Data Processing to Every Language}, 
      author={Guilherme Penedo and Hynek Kydlíček and Vinko Sabolčec and Bettina Messmer and Negar Foroutan and Amir Hossein Kargaran and Colin Raffel and Martin Jaggi and Leandro Von Werra and Thomas Wolf},
      year={2025},
      eprint={2506.20920},
      archivePrefix={arXiv},
      primaryClass={cs.CL},
      url={https://arxiv.org/abs/2506.20920}, 
}

@misc{eurollm9btech,
      title={EuroLLM-9B: Technical Report}, 
      author={Pedro Henrique Martins and João Alves and Patrick Fernandes and Nuno M. Guerreiro and Ricardo Rei and Amin Farajian and Mateusz Klimaszewski and Duarte M. Alves and José Pombal and Nicolas Boizard and Manuel Faysse and Pierre Colombo and François Yvon and Barry Haddow and José G. C. de Souza and Alexandra Birch and André F. T. Martins},
      year={2025},
      eprint={2506.04079},
      archivePrefix={arXiv},
      primaryClass={cs.CL},
      url={https://arxiv.org/abs/2506.04079}, 
}

@inproceedings{gao2025prolong,
  author       = {Wenhan Xiong and
                  Jingyu Liu and
                  Igor Molybog and
                  Hejia Zhang and
                  Prajjwal Bhargava and
                  Rui Hou and
                  Louis Martin and
                  Rashi Rungta and
                  Karthik Abinav Sankararaman and
                  Barlas Oguz and
                  Madian Khabsa and
                  Han Fang and
                  Yashar Mehdad and
                  Sharan Narang and
                  Kshitiz Malik and
                  Angela Fan and
                  Shruti Bhosale and
                  Sergey Edunov and
                  Mike Lewis and
                  Sinong Wang and
                  Hao Ma},
  title        = {Effective Long-Context Scaling of Foundation Models},
  booktitle    = {{NAACL} 2024},
  pages        = {4643--4663},
  publisher    = {ACL},
  year         = {2024},
  url          = {https://doi.org/10.18653/v1/2024.naacl-long.260},
  doi          = {10.18653/V1/2024.NAACL-LONG.260},
  bibsource    = {dblp computer science bibliography, https://dblp.org}
}

@article{frmt,
  author       = {Parker Riley and
                  Timothy Dozat and
                  Jan A. Botha and
                  Xavier Garcia and
                  Dan Garrette and
                  Jason Riesa and
                  Orhan Firat and
                  Noah Constant},
  title        = {{FRMT:} {A} Benchmark for Few-Shot Region-Aware Machine Translation},
  journal      = {Trans. Assoc. Comput. Linguistics},
  volume       = {11},
  pages        = {671--685},
  year         = {2023},
  url          = {https://doi.org/10.1162/tacl\_a\_00568},
  doi          = {10.1162/TACL\_A\_00568},
  bibsource    = {dblp computer science bibliography, https://dblp.org}
}

@article{Sousa2025,
  author    = {Hugo Sousa and Satya Almasian and Ricardo Campos and Alipio Jorge},
  title     = {Tradutor: Building a Variety Specific Translation Model},
  journal   = {AAAI},
  volume    = {39},
  number    = {24},
  pages     = {25183--25191},
  year      = {2025},
  doi       = {10.1609/aaai.v39i24.34704},
  issn      = {2374-3468},
  month     = {April}
}

@misc{bhardwaj2023redteaming,
      title={Red-Teaming Large Language Models using Chain of Utterances for Safety-Alignment}, 
      author={Rishabh Bhardwaj and Soujanya Poria},
      year={2023},
      eprint={2308.09662},
      archivePrefix={arXiv},
      primaryClass={cs.CL}
}

@article{cobbe2021gsm8k,
  author       = {Karl Cobbe and
                  Vineet Kosaraju and
                  Mohammad Bavarian and
                  Mark Chen and
                  Heewoo Jun and
                  Lukasz Kaiser and
                  Matthias Plappert and
                  Jerry Tworek and
                  Jacob Hilton and
                  Reiichiro Nakano and
                  Christopher Hesse and
                  John Schulman},
  title        = {Training Verifiers to Solve Math Word Problems},
  journal      = {CoRR},
  volume       = {abs/2110.14168},
  year         = {2021},
  url          = {https://arxiv.org/abs/2110.14168},
  eprinttype    = {arXiv},
  eprint       = {2110.14168},
  bibsource    = {dblp computer science bibliography, https://dblp.org}
}

@article{lozhkov2024starcoder2stackv2,
  author       = {Anton Lozhkov and
                  Raymond Li and
                  Loubna Ben Allal and
                  Federico Cassano and
                  Joel Lamy{-}Poirier and
                  Nouamane Tazi and
                  Ao Tang and
                  Dmytro Pykhtar and
                  Jiawei Liu and
                  Yuxiang Wei and
                  Tianyang Liu and
                  Max Tian and
                  Denis Kocetkov and
                  Arthur Zucker and
                  Younes Belkada and
                  Zijian Wang and
                  Qian Liu and
                  Dmitry Abulkhanov and
                  Indraneil Paul and
                  Zhuang Li and
                  Wen{-}Ding Li and
                  Megan Risdal and
                  Jia Li and
                  Jian Zhu and
                  Terry Yue Zhuo and
                  Evgenii Zheltonozhskii and
                  Nii Osae Osae Dade and
                  Wenhao Yu and
                  Lucas Krau{\ss} and
                  Naman Jain and
                  Yixuan Su and
                  Xuanli He and
                  Manan Dey and
                  Edoardo Abati and
                  Yekun Chai and
                  Niklas Muennighoff and
                  Xiangru Tang and
                  Muhtasham Oblokulov and
                  Christopher Akiki and
                  Marc Marone and
                  Chenghao Mou and
                  Mayank Mishra and
                  Alex Gu and
                  Binyuan Hui and
                  Tri Dao and
                  Armel Zebaze and
                  Olivier Dehaene and
                  Nicolas Patry and
                  Canwen Xu and
                  Julian J. McAuley and
                  Han Hu and
                  Torsten Scholak and
                  S{\'{e}}bastien Paquet and
                  Jennifer Robinson and
                  Carolyn Jane Anderson and
                  Nicolas Chapados and
                  et al.},
  title        = {StarCoder 2 and The Stack v2: The Next Generation},
  journal      = {CoRR},
  volume       = {abs/2402.19173},
  year         = {2024},
  url          = {https://doi.org/10.48550/arXiv.2402.19173},
  doi          = {10.48550/ARXIV.2402.19173},
  eprinttype    = {arXiv},
  eprint       = {2402.19173},
  bibsource    = {dblp computer science bibliography, https://dblp.org}
}

@misc{zhou2023ifeval,
      title={Instruction-Following Evaluation for Large Language Models}, 
      author={Jeffrey Zhou and Tianjian Lu and Swaroop Mishra and Siddhartha Brahma and Sujoy Basu and Yi Luan and Denny Zhou and Le Hou},
      year={2023},
      eprint={2311.07911},
      archivePrefix={arXiv},
      primaryClass={cs.CL},
      url={https://arxiv.org/abs/2311.07911}, 
}

@inproceedings{lin2022truthfulqa,
  title={Truthfulqa: Measuring how models mimic human falsehoods},
  author={Lin, Stephanie and Hilton, Jacob and Evans, Owain},
  booktitle={ACL 2022},
  pages={3214--3252},
  year={2022}
}

@article{srivastava2023bigbench,
  author       = {Aarohi Srivastava and
                  Abhinav Rastogi and
                  Abhishek Rao and
                  Abu Awal Md Shoeb and
                  Abubakar Abid and
                  Adam Fisch and
                  Adam R. Brown and
                  Adam Santoro and
                  Aditya Gupta and
                  Adri{\`{a}} Garriga{-}Alonso and
                  Agnieszka Kluska and
                  Aitor Lewkowycz and
                  Akshat Agarwal and
                  Alethea Power and
                  Alex Ray and
                  Alex Warstadt and
                  Alexander W. Kocurek and
                  Ali Safaya and
                  Ali Tazarv and
                  Alice Xiang and
                  Alicia Parrish and
                  Allen Nie and
                  Aman Hussain and
                  Amanda Askell and
                  Amanda Dsouza and
                  Ambrose Slone and
                  Ameet Rahane and
                  Anantharaman S. Iyer and
                  Anders Andreassen and
                  Andrea Madotto and
                  Andrea Santilli and
                  Andreas Stuhlm{\"{u}}ller and
                  Andrew M. Dai and
                  Andrew La and
                  Andrew K. Lampinen and
                  Andy Zou and
                  Angela Jiang and
                  Angelica Chen and
                  Anh Vuong and
                  Animesh Gupta and
                  Anna Gottardi and
                  Antonio Norelli and
                  Anu Venkatesh and
                  Arash Gholamidavoodi and
                  Arfa Tabassum and
                  Arul Menezes and
                  Arun Kirubarajan and
                  Asher Mullokandov and
                  Ashish Sabharwal and
                  Austin Herrick and
                  Avia Efrat and
                  Aykut Erdem and
                  Ayla Karakas and
                  B. Ryan Roberts and
                  Bao Sheng Loe and
                  Barret Zoph and
                  Bartlomiej Bojanowski and
                  Batuhan {\"{O}}zyurt and
                  Behnam Hedayatnia and
                  Behnam Neyshabur and
                  Benjamin Inden and
                  Benno Stein and
                  Berk Ekmekci and
                  Bill Yuchen Lin and
                  Blake Howald and
                  Bryan Orinion and
                  Cameron Diao and
                  Cameron Dour and
                  Catherine Stinson and
                  Cedrick Argueta and
                  C{\`{e}}sar Ferri Ram{\'{\i}}rez and
                  Chandan Singh and
                  Charles Rathkopf and
                  Chenlin Meng and
                  Chitta Baral and
                  Chiyu Wu and
                  Chris Callison{-}Burch and
                  Chris Waites and
                  Christian Voigt and
                  Christopher D. Manning and
                  Christopher Potts and
                  Cindy Ramirez and
                  Clara E. Rivera and
                  Clemencia Siro and
                  Colin Raffel and
                  Courtney Ashcraft and
                  Cristina Garbacea and
                  Damien Sileo and
                  Dan Garrette and
                  Dan Hendrycks and
                  Dan Kilman and
                  Dan Roth and
                  Daniel Freeman and
                  Daniel Khashabi and
                  Daniel Levy and
                  Daniel Mosegu{\'{\i}} Gonz{\'{a}}lez and
                  Danielle Perszyk and
                  Danny Hernandez and
                  Danqi Chen and
                  Daphne Ippolito and
                  Dar Gilboa and
                  David Dohan and
                  David Drakard and
                  David Jurgens and
                  Debajyoti Datta and
                  Deep Ganguli and
                  Denis Emelin and
                  Denis Kleyko and
                  Deniz Yuret and
                  Derek Chen and
                  Derek Tam and
                  Dieuwke Hupkes and
                  Diganta Misra and
                  Dilyar Buzan and
                  Dimitri Coelho Mollo and
                  Diyi Yang and
                  Dong{-}Ho Lee and
                  Dylan Schrader and
                  Ekaterina Shutova and
                  Ekin Dogus Cubuk and
                  Elad Segal and
                  Eleanor Hagerman and
                  Elizabeth Barnes and
                  Elizabeth Donoway and
                  Ellie Pavlick and
                  Emanuele Rodol{\`{a}} and
                  Emma Lam and
                  Eric Chu and
                  Eric Tang and
                  Erkut Erdem and
                  Ernie Chang and
                  Ethan A. Chi and
                  Ethan Dyer and
                  Ethan J. Jerzak and
                  Ethan Kim and
                  Eunice Engefu Manyasi and
                  Evgenii Zheltonozhskii and
                  Fanyue Xia and
                  Fatemeh Siar and
                  Fernando Mart{\'{\i}}nez{-}Plumed and
                  Francesca Happ{\'{e}} and
                  Fran{\c{c}}ois Chollet and
                  Frieda Rong and
                  Gaurav Mishra and
                  Genta Indra Winata and
                  Gerard de Melo and
                  Germ{\'{a}}n Kruszewski and
                  Giambattista Parascandolo and
                  Giorgio Mariani and
                  Gloria Wang and
                  Gonzalo Jaimovitch{-}L{\'{o}}pez and
                  Gregor Betz and
                  Guy Gur{-}Ari and
                  Hana Galijasevic and
                  Hannah Kim and
                  Hannah Rashkin and
                  Hannaneh Hajishirzi and
                  Harsh Mehta and
                  Hayden Bogar and
                  Henry Shevlin and
                  Hinrich Sch{\"{u}}tze and
                  Hiromu Yakura and
                  Hongming Zhang and
                  Hugh Mee Wong and
                  Ian Ng and
                  Isaac Noble and
                  Jaap Jumelet and
                  Jack Geissinger and
                  Jackson Kernion and
                  Jacob Hilton and
                  Jaehoon Lee and
                  Jaime Fern{\'{a}}ndez Fisac and
                  James B. Simon and
                  James Koppel and
                  James Zheng and
                  James Zou and
                  Jan Kocon and
                  Jana Thompson and
                  Janelle Wingfield and
                  Jared Kaplan and
                  Jarema Radom and
                  Jascha Sohl{-}Dickstein and
                  Jason Phang and
                  Jason Wei and
                  Jason Yosinski and
                  Jekaterina Novikova and
                  Jelle Bosscher and
                  Jennifer Marsh and
                  Jeremy Kim and
                  Jeroen Taal and
                  Jesse H. Engel and
                  Jesujoba Alabi and
                  Jiacheng Xu and
                  Jiaming Song and
                  Jillian Tang and
                  Joan Waweru and
                  John Burden and
                  John Miller and
                  John U. Balis and
                  Jonathan Batchelder and
                  Jonathan Berant and
                  J{\"{o}}rg Frohberg and
                  Jos Rozen and
                  Jos{\'{e}} Hern{\'{a}}ndez{-}Orallo and
                  Joseph Boudeman and
                  Joseph Guerr and
                  Joseph Jones and
                  Joshua B. Tenenbaum and
                  Joshua S. Rule and
                  Joyce Chua and
                  Kamil Kanclerz and
                  Karen Livescu and
                  Karl Krauth and
                  Karthik Gopalakrishnan and
                  Katerina Ignatyeva and
                  Katja Markert and
                  Kaustubh D. Dhole and
                  Kevin Gimpel and
                  Kevin Omondi and
                  Kory W. Mathewson and
                  Kristen Chiafullo and
                  Ksenia Shkaruta and
                  Kumar Shridhar and
                  Kyle McDonell and
                  Kyle Richardson and
                  Laria Reynolds and
                  Leo Gao and
                  Li Zhang and
                  Liam Dugan and
                  Lianhui Qin and
                  Lidia Contreras Ochando and
                  Louis{-}Philippe Morency and
                  Luca Moschella and
                  Lucas Lam and
                  Lucy Noble and
                  Ludwig Schmidt and
                  Luheng He and
                  Luis Oliveros Col{\'{o}}n and
                  Luke Metz and
                  L{\"{u}}tfi Kerem Senel and
                  Maarten Bosma and
                  Maarten Sap and
                  Maartje ter Hoeve and
                  Maheen Farooqi and
                  Manaal Faruqui and
                  Mantas Mazeika and
                  Marco Baturan and
                  Marco Marelli and
                  Marco Maru and
                  Mar{\'{\i}}a Jos{\'{e}} Ram{\'{\i}}rez{-}Quintana and
                  Marie Tolkiehn and
                  Mario Giulianelli and
                  Martha Lewis and
                  Martin Potthast and
                  Matthew L. Leavitt and
                  Matthias Hagen and
                  M{\'{a}}ty{\'{a}}s Schubert and
                  Medina Baitemirova and
                  Melody Arnaud and
                  Melvin McElrath and
                  Michael A. Yee and
                  Michael Cohen and
                  Michael Gu and
                  Michael I. Ivanitskiy and
                  Michael Starritt and
                  Michael Strube and
                  Michal Swedrowski and
                  Michele Bevilacqua and
                  Michihiro Yasunaga and
                  Mihir Kale and
                  Mike Cain and
                  Mimee Xu and
                  Mirac Suzgun and
                  Mitch Walker and
                  Mo Tiwari and
                  Mohit Bansal and
                  Moin Aminnaseri and
                  Mor Geva and
                  Mozhdeh Gheini and
                  Mukund Varma T. and
                  Nanyun Peng and
                  Nathan A. Chi and
                  Nayeon Lee and
                  Neta Gur{-}Ari Krakover and
                  Nicholas Cameron and
                  Nicholas Roberts and
                  Nick Doiron and
                  Nicole Martinez and
                  Nikita Nangia and
                  Niklas Deckers and
                  Niklas Muennighoff and
                  Nitish Shirish Keskar and
                  Niveditha Iyer and
                  Noah Constant and
                  Noah Fiedel and
                  Nuan Wen and
                  Oliver Zhang and
                  Omar Agha and
                  Omar Elbaghdadi and
                  Omer Levy and
                  Owain Evans and
                  Pablo Antonio Moreno Casares and
                  Parth Doshi and
                  Pascale Fung and
                  Paul Pu Liang and
                  Paul Vicol and
                  Pegah Alipoormolabashi and
                  Peiyuan Liao and
                  Percy Liang and
                  Peter Chang and
                  Peter Eckersley and
                  Phu Mon Htut and
                  Pinyu Hwang and
                  Piotr Milkowski and
                  Piyush Patil and
                  Pouya Pezeshkpour and
                  Priti Oli and
                  Qiaozhu Mei and
                  Qing Lyu and
                  Qinlang Chen and
                  Rabin Banjade and
                  Rachel Etta Rudolph and
                  Raefer Gabriel and
                  Rahel Habacker and
                  Ramon Risco and
                  Rapha{\"{e}}l Milli{\`{e}}re and
                  Rhythm Garg and
                  Richard Barnes and
                  Rif A. Saurous and
                  Riku Arakawa and
                  Robbe Raymaekers and
                  Robert Frank and
                  Rohan Sikand and
                  Roman Novak and
                  Roman Sitelew and
                  Ronan LeBras and
                  Rosanne Liu and
                  Rowan Jacobs and
                  Rui Zhang and
                  Ruslan Salakhutdinov and
                  Ryan Chi and
                  Ryan Lee and
                  Ryan Stovall and
                  Ryan Teehan and
                  Rylan Yang and
                  Sahib Singh and
                  Saif M. Mohammad and
                  Sajant Anand and
                  Sam Dillavou and
                  Sam Shleifer and
                  Sam Wiseman and
                  Samuel Gruetter and
                  Samuel R. Bowman and
                  Samuel S. Schoenholz and
                  Sanghyun Han and
                  Sanjeev Kwatra and
                  Sarah A. Rous and
                  Sarik Ghazarian and
                  Sayan Ghosh and
                  Sean Casey and
                  Sebastian Bischoff and
                  Sebastian Gehrmann and
                  Sebastian Schuster and
                  Sepideh Sadeghi and
                  Shadi Hamdan and
                  Sharon Zhou and
                  Shashank Srivastava and
                  Sherry Shi and
                  Shikhar Singh and
                  Shima Asaadi and
                  Shixiang Shane Gu and
                  Shubh Pachchigar and
                  Shubham Toshniwal and
                  Shyam Upadhyay and
                  Shyamolima (Shammie) Debnath and
                  Siamak Shakeri and
                  Simon Thormeyer and
                  Simone Melzi and
                  Siva Reddy and
                  Sneha Priscilla Makini and
                  Soo{-}Hwan Lee and
                  Spencer Torene and
                  Sriharsha Hatwar and
                  Stanislas Dehaene and
                  Stefan Divic and
                  Stefano Ermon and
                  Stella Biderman and
                  Stephanie Lin and
                  Stephen Prasad and
                  Steven T. Piantadosi and
                  Stuart M. Shieber and
                  Summer Misherghi and
                  Svetlana Kiritchenko and
                  Swaroop Mishra and
                  Tal Linzen and
                  Tal Schuster and
                  Tao Li and
                  Tao Yu and
                  Tariq Ali and
                  Tatsu Hashimoto and
                  Te{-}Lin Wu and
                  Th{\'{e}}o Desbordes and
                  Theodore Rothschild and
                  Thomas Phan and
                  Tianle Wang and
                  Tiberius Nkinyili and
                  Timo Schick and
                  Timofei Kornev and
                  Titus Tunduny and
                  Tobias Gerstenberg and
                  Trenton Chang and
                  Trishala Neeraj and
                  Tushar Khot and
                  Tyler Shultz and
                  Uri Shaham and
                  Vedant Misra and
                  Vera Demberg and
                  Victoria Nyamai and
                  Vikas Raunak and
                  Vinay V. Ramasesh and
                  Vinay Uday Prabhu and
                  Vishakh Padmakumar and
                  Vivek Srikumar and
                  William Fedus and
                  William Saunders and
                  William Zhang and
                  Wout Vossen and
                  Xiang Ren and
                  Xiaoyu Tong and
                  Xinran Zhao and
                  Xinyi Wu and
                  Xudong Shen and
                  Yadollah Yaghoobzadeh and
                  Yair Lakretz and
                  Yangqiu Song and
                  Yasaman Bahri and
                  Yejin Choi and
                  Yichi Yang and
                  Yiding Hao and
                  Yifu Chen and
                  Yonatan Belinkov and
                  Yu Hou and
                  Yufang Hou and
                  Yuntao Bai and
                  Zachary Seid and
                  Zhuoye Zhao and
                  Zijian Wang and
                  Zijie J. Wang and
                  Zirui Wang and
                  Ziyi Wu},
  title        = {Beyond the Imitation Game: Quantifying and extrapolating the capabilities
                  of language models},
  journal      = {Trans. Mach. Learn. Res.},
  volume       = {2023},
  year         = {2023},
  url          = {https://openreview.net/forum?id=uyTL5Bvosj},
  bibsource    = {dblp computer science bibliography, https://dblp.org}
}

@article{vidgen2023simplesafetytests,
  author       = {Bertie Vidgen and
                  Hannah Rose Kirk and
                  Rebecca Qian and
                  Nino Scherrer and
                  Anand Kannappan and
                  Scott A. Hale and
                  Paul R{\"{o}}ttger},
  title        = {SimpleSafetyTests: a Test Suite for Identifying Critical Safety Risks
                  in Large Language Models},
  journal      = {CoRR},
  volume       = {abs/2311.08370},
  year         = {2023},
  url          = {https://doi.org/10.48550/arXiv.2311.08370},
  doi          = {10.48550/ARXIV.2311.08370},
  eprinttype    = {arXiv},
  eprint       = {2311.08370},
  bibsource    = {dblp computer science bibliography, https://dblp.org}
}

@inproceedings{rottger2024xstest,
  author       = {Paul R{\"{o}}ttger and
                  Hannah Kirk and
                  Bertie Vidgen and
                  Giuseppe Attanasio and
                  Federico Bianchi and
                  Dirk Hovy},
  title        = {XSTest: {A} Test Suite for Identifying Exaggerated Safety Behaviours
                  in Large Language Models},
  booktitle    = {{NAACL} 2024 (Volume 1: Long Papers)},
  pages        = {5377--5400},
  publisher    = {ACL},
  year         = {2024},
  url          = {https://doi.org/10.18653/v1/2024.naacl-long.301},
  doi          = {10.18653/V1/2024.NAACL-LONG.301},
  bibsource    = {dblp computer science bibliography, https://dblp.org}
}

@inproceedings{gloria_llm,
  author       = {Ricardo Lopes and
                  Jo{\~{a}}o Magalh{\~{a}}es and
                  David Semedo},
  title        = {Gl{\'{o}}rIA: {A} Generative and Open Large Language Model for
                  Portuguese},
  booktitle    = {{PROPOR} 2024, Volume 1},
  pages        = {441--453},
  publisher    = {ACL},
  year         = {2024},
  url          = {https://aclanthology.org/2024.propor-1.45},
  bibsource    = {dblp computer science bibliography, https://dblp.org}
}

@article{gemma_2025,
  title={Gemma 3 technical report},
  author={Kamath, Aishwarya and Ferret, Johan and Pathak, Shreya and Vieillard, Nino and Merhej, Ramona and Perrin, Sarah and Matejovicova, Tatiana and Ram{\'e}, Alexandre and Rivi{\`e}re, Morgane and others},
  journal={arXiv preprint arXiv:2503.19786},
  url={https://arxiv.org/abs/2503.19786},
  year={2025}
}

@misc{rafailov2024directpreferenceoptimizationlanguage,
      title={Direct Preference Optimization: Your Language Model is Secretly a Reward Model}, 
      author={Rafael Rafailov and Archit Sharma and Eric Mitchell and Stefano Ermon and Christopher D. Manning and Chelsea Finn},
      year={2024},
      eprint={2305.18290},
      archivePrefix={arXiv},
      primaryClass={cs.LG},
      url={https://arxiv.org/abs/2305.18290}, 
}

@misc{liu2023what,
      title={What Makes Good Data for Alignment? A Comprehensive Study of Automatic Data Selection in Instruction Tuning}, 
      author={Wei Liu and Weihao Zeng and Keqing He and Yong Jiang and Junxian He},
      year={2023},
      eprint={2312.15685},
      archivePrefix={arXiv},
      primaryClass={cs.CL}
}

@article{apertus,
  author       = {Project Apertus},
  title        = {Apertus: Democratizing Open and Compliant LLMs for Global Language
                  Environments},
  journal      = {CoRR},
  volume       = {abs/2509.14233},
  year         = {2025},
  url          = {https://doi.org/10.48550/arXiv.2509.14233},
  doi          = {10.48550/ARXIV.2509.14233},
  eprinttype    = {arXiv},
  eprint       = {2509.14233},
  bibsource    = {dblp computer science bibliography, https://dblp.org}
}

@article{gemma2,
  author       = {Morgane Rivi{\`{e}}re and
                  Shreya Pathak and
                  Pier Giuseppe Sessa and
                  Cassidy Hardin and
                  Surya Bhupatiraju and
                  L{\'{e}}onard Hussenot and
                  Thomas Mesnard and
                  Bobak Shahriari and
                  Alexandre Ram{\'{e}} and
                  Johan Ferret and
                  Peter Liu and
                  Pouya Tafti and
                  Abe Friesen and
                  Michelle Casbon and
                  Sabela Ramos and
                  Ravin Kumar and
                  Charline Le Lan and
                  Sammy Jerome and
                  Anton Tsitsulin and
                  Nino Vieillard and
                  Piotr Stanczyk and
                  Sertan Girgin and
                  Nikola Momchev and
                  Matt Hoffman and
                  Shantanu Thakoor and
                  Jean{-}Bastien Grill and
                  Behnam Neyshabur and
                  Olivier Bachem and
                  Alanna Walton and
                  Aliaksei Severyn and
                  Alicia Parrish and
                  Aliya Ahmad and
                  Allen Hutchison and
                  Alvin Abdagic and
                  Amanda Carl and
                  Amy Shen and
                  Andy Brock and
                  Andy Coenen and
                  Anthony Laforge and
                  Antonia Paterson and
                  Ben Bastian and
                  Bilal Piot and
                  Bo Wu and
                  Brandon Royal and
                  Charlie Chen and
                  Chintu Kumar and
                  Chris Perry and
                  Chris Welty and
                  Christopher A. Choquette{-}Choo and
                  Danila Sinopalnikov and
                  David Weinberger and
                  Dimple Vijaykumar and
                  Dominika Rogozinska and
                  Dustin Herbison and
                  Elisa Bandy and
                  Emma Wang and
                  Eric Noland and
                  Erica Moreira and
                  Evan Senter and
                  Evgenii Eltyshev and
                  Francesco Visin and
                  Gabriel Rasskin and
                  Gary Wei and
                  Glenn Cameron and
                  Gus Martins and
                  Hadi Hashemi and
                  Hanna Klimczak{-}Plucinska and
                  Harleen Batra and
                  Harsh Dhand and
                  Ivan Nardini and
                  Jacinda Mein and
                  Jack Zhou and
                  James Svensson and
                  Jeff Stanway and
                  Jetha Chan and
                  Jin Peng Zhou and
                  Joana Carrasqueira and
                  Joana Iljazi and
                  Jocelyn Becker and
                  Joe Fernandez and
                  Joost van Amersfoort and
                  Josh Gordon and
                  Josh Lipschultz and
                  Josh Newlan and
                  Ju{-}yeong Ji and
                  Kareem Mohamed and
                  Kartikeya Badola and
                  Kat Black and
                  Katie Millican and
                  Keelin McDonell and
                  Kelvin Nguyen and
                  Kiranbir Sodhia and
                  Kish Greene and
                  Lars Lowe Sj{\"{o}}sund and
                  Lauren Usui and
                  Laurent Sifre and
                  Lena Heuermann and
                  Leticia Lago and
                  Lilly McNealus},
  title        = {Gemma 2: Improving Open Language Models at a Practical Size},
  journal      = {CoRR},
  volume       = {abs/2408.00118},
  year         = {2024},
  url          = {https://doi.org/10.48550/arXiv.2408.00118},
  doi          = {10.48550/ARXIV.2408.00118},
  eprinttype    = {arXiv},
  eprint       = {2408.00118},
  bibsource    = {dblp computer science bibliography, https://dblp.org}
}

@article{mistral,
  author       = {Albert Q. Jiang and
                  Alexandre Sablayrolles and
                  Arthur Mensch and
                  Chris Bamford and
                  Devendra Singh Chaplot and
                  Diego de Las Casas and
                  Florian Bressand and
                  Gianna Lengyel and
                  Guillaume Lample and
                  Lucile Saulnier and
                  L{\'{e}}lio Renard Lavaud and
                  Marie{-}Anne Lachaux and
                  Pierre Stock and
                  Teven Le Scao and
                  Thibaut Lavril and
                  Thomas Wang and
                  Timoth{\'{e}}e Lacroix and
                  William El Sayed},
  title        = {Mistral 7B},
  journal      = {CoRR},
  volume       = {abs/2310.06825},
  year         = {2023},
  url          = {https://doi.org/10.48550/arXiv.2310.06825},
  doi          = {10.48550/ARXIV.2310.06825},
  eprinttype    = {arXiv},
  eprint       = {2310.06825},
  bibsource    = {dblp computer science bibliography, https://dblp.org}
}

@article{qwen2.5,
  author       = {An Yang and
                  Baosong Yang and
                  Beichen Zhang and
                  Binyuan Hui and
                  Bo Zheng and
                  Bowen Yu and
                  Chengyuan Li and
                  Dayiheng Liu and
                  Fei Huang and
                  Haoran Wei and
                  Huan Lin and
                  Jian Yang and
                  Jianhong Tu and
                  Jianwei Zhang and
                  Jianxin Yang and
                  Jiaxi Yang and
                  Jingren Zhou and
                  Junyang Lin and
                  Kai Dang and
                  Keming Lu and
                  Keqin Bao and
                  Kexin Yang and
                  Le Yu and
                  Mei Li and
                  Mingfeng Xue and
                  Pei Zhang and
                  Qin Zhu and
                  Rui Men and
                  Runji Lin and
                  Tianhao Li and
                  Tingyu Xia and
                  Xingzhang Ren and
                  Xuancheng Ren and
                  Yang Fan and
                  Yang Su and
                  Yichang Zhang and
                  Yu Wan and
                  Yuqiong Liu and
                  Zeyu Cui and
                  Zhenru Zhang and
                  Zihan Qiu},
  title        = {Qwen2.5 Technical Report},
  journal      = {CoRR},
  volume       = {abs/2412.15115},
  year         = {2024},
  url          = {https://doi.org/10.48550/arXiv.2412.15115},
  doi          = {10.48550/ARXIV.2412.15115},
  eprinttype    = {arXiv},
  eprint       = {2412.15115},
  bibsource    = {dblp computer science bibliography, https://dblp.org}
}

@inproceedings{vllm,
  author       = {Woosuk Kwon and
                  Zhuohan Li and
                  Siyuan Zhuang and
                  Ying Sheng and
                  Lianmin Zheng and
                  Cody Hao Yu and
                  Joseph Gonzalez and
                  Hao Zhang and
                  Ion Stoica},
  title        = {Efficient Memory Management for Large Language Model Serving with
                  PagedAttention},
  booktitle    = {{SOSP} 2023},
  pages        = {611--626},
  publisher    = {{ACM}},
  year         = {2023},
  url          = {https://doi.org/10.1145/3600006.3613165},
  doi          = {10.1145/3600006.3613165},
  bibsource    = {dblp computer science bibliography, https://dblp.org}
}

@article{arc_c,
  author       = {Peter Clark and
                  Isaac Cowhey and
                  Oren Etzioni and
                  Tushar Khot and
                  Ashish Sabharwal and
                  Carissa Schoenick and
                  Oyvind Tafjord},
  title        = {Think you have Solved Question Answering? Try ARC, the {AI2} Reasoning
                  Challenge},
  journal      = {CoRR},
  volume       = {abs/1803.05457},
  year         = {2018},
  url          = {http://arxiv.org/abs/1803.05457},
  eprinttype    = {arXiv},
  eprint       = {1803.05457},
  bibsource    = {dblp computer science bibliography, https://dblp.org}
}

@inproceedings{mmlu,
  author       = {Dan Hendrycks and
                  Collin Burns and
                  Steven Basart and
                  Andy Zou and
                  Mantas Mazeika and
                  Dawn Song and
                  Jacob Steinhardt},
  title        = {Measuring Massive Multitask Language Understanding},
  booktitle    = {9th International Conference on Learning Representations, {ICLR} 2021,
                  Virtual Event, Austria, May 3-7, 2021},
  publisher    = {OpenReview.net},
  year         = {2021},
  url          = {https://openreview.net/forum?id=d7KBjmI3GmQ},
  bibsource    = {dblp computer science bibliography, https://dblp.org}
}

@inproceedings{piqa,
  author       = {Yonatan Bisk and
                  Rowan Zellers and
                  Ronan Le Bras and
                  Jianfeng Gao and
                  Yejin Choi},
  title        = {{PIQA:} Reasoning about Physical Commonsense in Natural Language},
  booktitle    = {{AAAI} 2020},
  pages        = {7432--7439},
  publisher    = {{AAAI} Press},
  year         = {2020},
  url          = {https://doi.org/10.1609/aaai.v34i05.6239},
  doi          = {10.1609/AAAI.V34I05.6239},
  bibsource    = {dblp computer science bibliography, https://dblp.org}
}

@misc{nvidia/Nemotron-Personas-USA,
  author = {Meyer, Yev and Corneil, Dane},
  title = {{Nemotron-Personas-USA}: Synthetic Personas Aligned to Real-World Distributions},
  month = {June},
  year = {2025},
  url = {https://huggingface.co/datasets/nvidia/Nemotron-Personas-USA}
}

@misc{deutsch2025wmt24expandinglanguagecoverage,
      title={WMT24++: Expanding the Language Coverage of WMT24 to 55 Languages \& Dialects}, 
      author={Daniel Deutsch and Eleftheria Briakou and Isaac Caswell and Mara Finkelstein and Rebecca Galor and Juraj Juraska and Geza Kovacs and Alison Lui and Ricardo Rei and Jason Riesa and Shruti Rijhwani and Parker Riley and Elizabeth Salesky and Firas Trabelsi and Stephanie Winkler and Biao Zhang and Markus Freitag},
      year={2025},
      eprint={2502.12404},
      archivePrefix={arXiv},
      primaryClass={cs.CL},
      url={https://arxiv.org/abs/2502.12404}, 
}

@article{qwen_guard_3,
  author       = {Haiquan Zhao and
                  Chenhan Yuan and
                  Fei Huang and
                  Xiaomeng Hu and
                  Yichang Zhang and
                  An Yang and
                  Bowen Yu and
                  Dayiheng Liu and
                  Jingren Zhou and
                  Junyang Lin and
                  Baosong Yang and
                  Chen Cheng and
                  Jialong Tang and
                  Jiandong Jiang and
                  Jianwei Zhang and
                  Jijie Xu and
                  Ming Yan and
                  Minmin Sun and
                  Pei Zhang and
                  Pengjun Xie and
                  Qiaoyu Tang and
                  Qin Zhu and
                  Rong Zhang and
                  Shibin Wu and
                  Shuo Zhang and
                  Tao He and
                  Tianyi Tang and
                  Tingyu Xia and
                  Wei Liao and
                  Weizhou Shen and
                  Wenbiao Yin and
                  Wenmeng Zhou and
                  Wenyuan Yu and
                  Xiaobin Wang and
                  Xiaodong Deng and
                  Xiaodong Xu and
                  Xinyu Zhang and
                  Yang Liu and
                  Yeqiu Li and
                  Yi Zhang and
                  Yong Jiang and
                  Yu Wan and
                  Yuxin Zhou},
  title        = {Qwen3Guard Technical Report},
  journal      = {CoRR},
  volume       = {abs/2510.14276},
  year         = {2025},
  url          = {https://doi.org/10.48550/arXiv.2510.14276},
  doi          = {10.48550/ARXIV.2510.14276},
  eprinttype    = {arXiv},
  eprint       = {2510.14276},
  bibsource    = {dblp computer science bibliography, https://dblp.org}
}

@inproceedings{garciagasulla2025efficientsafetyretrofittingjailbreaking,
  author       = {Dario Garcia{-}Gasulla and
                  Adri{\'{a}}n Tormos and
                  Anna Arias{-}Duart and
                  Daniel Hinjos and
                  Oscar Molina{-}Sedano and
                  Ashwin Kumar Gurarajan and
                  Maria Eugenia Cardello},
  editor       = {Martin T{\"{o}}rngren and
                  Barbara Gallina and
                  Erwin Schoitsch and
                  Elena Troubitsyna and
                  Friedemann Bitsch},
  title        = {Efficient Safety Retrofitting Against Jailbreaking for LLMs},
  booktitle    = {{SAFECOMP} 2025 Workshops},
  series       = {Lecture Notes in Computer Science},
  volume       = {15955},
  pages        = {537--565},
  publisher    = {Springer},
  year         = {2025},
  url          = {https://doi.org/10.1007/978-3-032-02018-5\_39},
  doi          = {10.1007/978-3-032-02018-5\_39},
  bibsource    = {dblp computer science bibliography, https://dblp.org}
}

@article{siqa,
  author       = {Maarten Sap and
                  Hannah Rashkin and
                  Derek Chen and
                  Ronan Le Bras and
                  Yejin Choi},
  title        = {SocialIQA: Commonsense Reasoning about Social Interactions},
  journal      = {CoRR},
  volume       = {abs/1904.09728},
  year         = {2019},
  url          = {http://arxiv.org/abs/1904.09728},
  eprinttype    = {arXiv},
  eprint       = {1904.09728},
  bibsource    = {dblp computer science bibliography, https://dblp.org}
}

@article{advbench,
  author       = {Andy Zou and
                  Zifan Wang and
                  J. Zico Kolter and
                  Matt Fredrikson},
  title        = {Universal and Transferable Adversarial Attacks on Aligned Language
                  Models},
  journal      = {CoRR},
  volume       = {abs/2307.15043},
  year         = {2023},
  url          = {https://doi.org/10.48550/arXiv.2307.15043},
  doi          = {10.48550/ARXIV.2307.15043},
  eprinttype    = {arXiv},
  eprint       = {2307.15043},
  bibsource    = {dblp computer science bibliography, https://dblp.org}
}

@misc{deepseekai2024deepseekv32,
      title={DeepSeek-V3.2-Exp: Boosting Long-Context Efficiency with DeepSeek Sparse Attention}, 
      author={DeepSeek-AI},
      url={https://github.com/deepseek-ai/DeepSeek-V3.2-Exp/blob/main/DeepSeek_V3_2.pdf},
      year={2025},
}

@misc{NemotronPostTrainingDatasetV1,
      author = {Nathawani, Dhruv and Gitman, Igor and Majumdar, Somshubra and Bakhturina, Evelina and Sunil Mahabaleshwarkar, Ameya and and Zhang, Jian and Polak Scowcroft, Jane},
      title = {{Nemotron-Post-Training-Dataset-v1}},
      version = {1.0},
      publisher = {{NVIDIA}},
      year = {2025},
      url = {https://huggingface.co/datasets/nvidia/Nemotron-Post-Training-Dataset-v1}
}

@misc{NemotronPostTrainingDatasetV2,
      author = {Nathawani, Dhruv and Ding, Shuoyang and Lavrukhin, Vitaly and Gitman, Igor and Majumdar, Somshubra and Bakhturina, Evelina and Ginsburg, Boris and Polak Scowcroft, Jane},
      title = {{Nemotron-Post-Training-Dataset-v2}},
      version = {2.0},
      publisher = {{NVIDIA}},
      year = {2025}, month = aug,
      url = {https://huggingface.co/datasets/nvidia/Nemotron-Post-Training-Dataset-v2}
}

@misc{nvidia2025nvidianemotronnano2,
      title={NVIDIA Nemotron Nano 2: An Accurate and Efficient Hybrid Mamba-Transformer Reasoning Model},
      author={NVIDIA},
      year={2025},
      eprint={2508.14444},
      archivePrefix={arXiv},
      primaryClass={cs.CL},
      url={https://arxiv.org/abs/2508.14444},
}

\end{document}